%% file: aaai24.tex
\documentclass[letterpaper]{article} 
\usepackage{aaai2026} 
\usepackage[table]{xcolor}
\usepackage{times}  
\usepackage{helvet}  
\usepackage{courier}  
\usepackage[hyphens]{url}  
\usepackage{graphicx} 
\usepackage{natbib}  
\usepackage{caption} 
\usepackage{algorithm}
\usepackage{algorithmic}
\usepackage{booktabs}
\usepackage{tabularray}
\usepackage{multirow}
\usepackage{graphicx}      
\usepackage{booktabs}      
\usepackage[table]{xcolor} 

\usepackage[misc]{ifsym}
\usepackage{twemojis}
\usepackage{url}
\usepackage{amsmath,amssymb}

\usepackage{amsfonts}
\usepackage{helvet}
\usepackage{courier}
\usepackage{url}
\usepackage{float,color}
\usepackage{times}
\usepackage{algorithm}
\usepackage{algorithmic}
\usepackage{float}
\usepackage{pdfpages}
\usepackage{epsfig}
\usepackage{url}
\usepackage{diagbox}
\usepackage{subfigure}
\usepackage{epstopdf}
\usepackage{multicol}
\usepackage{makecell}
\usepackage{longtable}
\usepackage{dsfont,amsfonts,color}
\usepackage{multirow}
\usepackage{booktabs}
\usepackage{xr}
\usepackage{tablefootnote}
\def\0{{\bf 0}}
\def\1{{\bf 1}}

\usepackage{newfloat}
\usepackage{listings}
\DeclareCaptionStyle{ruled}{labelfont=normalfont,labelsep=colon,strut=off} 
\floatstyle{ruled}
\newfloat{listing}{tb}{lst}{}
\floatname{listing}{Listing}
\title{RCP-Merging: Merging Long Chain-of-Thought Models with Domain-Specific Models by Considering Reasoning Capability as Prior}
\author{
    Junyao Yang\textsuperscript{\rm 1},
    Jianwei Wang\textsuperscript{\rm 1},
    Huiping Zhuang\textsuperscript{\rm 1},
    Cen Chen\textsuperscript{\rm 1},
    Ziqian Zeng\textsuperscript{\rm 1 \thanks{Corresponding author.}} 
}
\affiliations{
    \textsuperscript{\rm 1}South China University of Technology, China\\

}

\usepackage{bibentry}
\usepackage[switch]{lineno} 
\begin{document}
\maketitle

\input{abstract}
\input{introduction}
\input{related_work}

\input{Preliminary}
\input{methodology}

\input{experiment}
\subsection{Hypterparameter Analysis.}
\input{appendix/hyperparameters}
\input{ablation}

\input{conclusion}

\section{Acknowledgement}
This work was supported by National Natural Science Foundation of China (62406114,62472181,62306117), the Fundamental Research Funds for the Central Universities (2024ZYGXZR074), Guangdong Basic and Applied Basic Research Foundation (2025A1515011413,2024A04J3681,2024A1515010220), and GJYC program of Guangzhou (2024D03J0005), National Key R \& D Project from Minister of Science and Technology (2024YFA1211500), and CCF-Baidu Open Fund.

\small \bibliography{aaai24}
\input{checklist}
\input{appendix}

\end{document}

%% file: abstract.tex

\begin{abstract}
Large Language Models (LLMs) with long chain-of-thought (CoT) capability, termed Reasoning Models, demonstrate superior intricate problem-solving abilities through multi-step long CoT reasoning.
To create a dual-capability model with long CoT capability and domain-specific knowledge without substantial computational and data costs, model merging emerges as a highly resource-efficient method. 
However, significant challenges lie in merging domain-specific LLMs with long CoT ones since nowadays merging methods suffer from reasoning capability degradation, even gibberish output and output collapse. 
To overcome this, we introduce \textbf{RCP-Merging}:
Merging Long Chain-of-Thought Models with Domain-Specific Models by Considering \textbf{R}easoning \textbf{C}apability as \textbf{P}rior, 
a novel merging framework designed to integrate domain-specific LLMs with long CoT capability, meanwhile maintaining model performance in the original domain. 
Treating reasoning model weights as foundational prior, our method utilizes a reasoning capability indicator to preserve core long CoT capability model weights while selectively merging essential domain-specific weights. 
We conducted extensive experiments on Qwen2.5-7B, Llama3.1-8B, and Qwen2.5-1.5B models in BioMedicine and Finance domains.
Our results show that RCP-Merging successfully merges a reasoning model with domain-specific ones, improving domain task performance by 9.5\% and 9.2\% over state-of-the-art methods, without significantly harming the original long CoT reasoning capability.


\end{abstract}

\begin{links}
    \link{Code}{https://github.com/ZeroNLP/RCP-Merging}
    \link{Datasets}{https://github.com/ZeroNLP/RCP-Merging}
\end{links}

%% file: introduction.tex
\section*{Introduction}

Large Language Models (LLMs) with long chain-of-thought (CoT) capability, termed Reasoning Models, have demonstrated exceptional performance on complex reasoning tasks 
\citep{jaech2024openai,openai2025o3,guo2025deepseek,xai2025grok35}.
Mostly trained on verifiable tasks like code generation and mathematical reasoning, the results in Table \ref{tab:performance_comparison} show that the reasoning model demonstrates relatively weak performance compared with models that specifically fine-tune on a certain domain. 
However, long CoT's multi-step reasoning deduction is critical for complex problem-solving in specific domains like BioMedicine and Finance, extending beyond simple information retrieval \citep{cui2025curie,Tang2025generallongcot}. 
Moreover, the scarcity of models specifically trained for these fields remains a key challenge.
%
This difficulty stems from current long CoT realization relying on additional training, which introduces challenges like catastrophic forgetting, inefficient resource allocation, not to mention the inherent difficulty in obtaining high-quality domain reasoning data \citep{dong25LongReD,zhang2025logit,zeng25revisiting}.

\begin{figure}[t]
\centering

\includegraphics[width=0.37\textwidth]{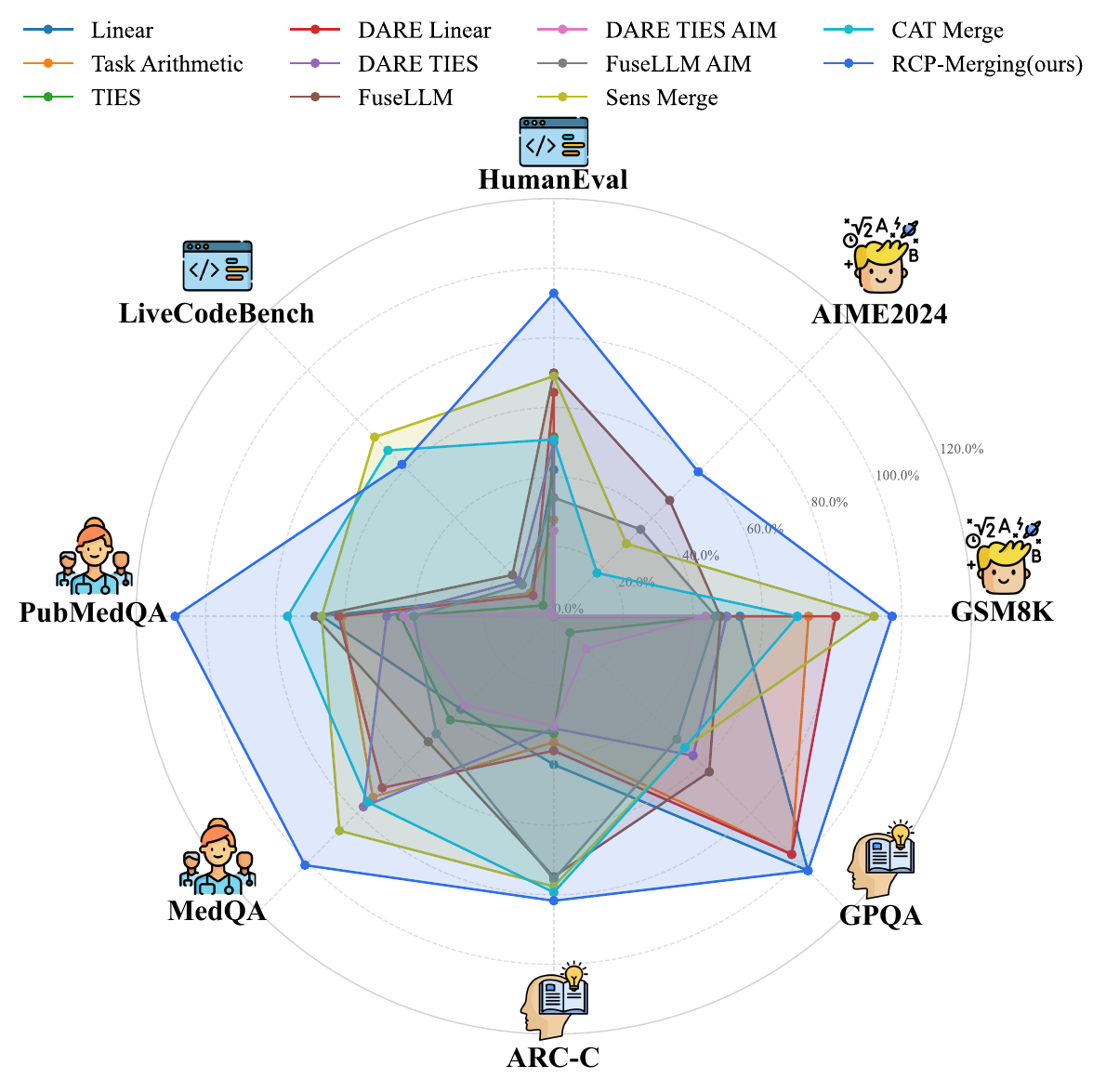}
\caption{Performance comparison of RCP-Merging and other methods in merging Qwen2.5-7B, Meditron3-Qwen2.5-7B, and DeepSeek-R1-Distill-Qwen-7B on eight datasets in Math, Code, BioMedicine, and Knowledge areas.
}
\label{fig:radar}

\end{figure}

Fortunately, model merging \citep{li2023deepmodelfusion,ilharco2023editing,yang2024modelmerge} has recently  emerged as a resource-efficient technique to create a single model with dual capabilities without requiring extra training data. 
However, a significant gap exists that current model merging focuses on combining models for certain domains, such as merging a model specialized in General Knowledge with one for Chinese. 
As results show in the LiveCodeBench \citep{jain2024livecodebench} and AIME \citep{aime_1983_2024} datasets in Figure \ref{fig:radar}, trying to merge a reasoning model with a domain-specific one often leads to a collapse of the output and a sharp performance decline.
Therefore, it is highly valuable to find a method that can successfully integrate a domain-specific model with a reasoning model and subsequently boost the merged model's performance on its original domain-specific tasks. 

To tackle this problem, existing merging methods often struggle to preserve long CoT capabilities when integrating reasoning models with domain-specific ones. 
For instance, some methods \citep{ilharco2023editing,wan24fusellm} operate under the assumption that larger weights are more important. 
By trimming the smaller weights \citep{yadav2023tiesmerging} or rescaling the larger weights \citep{yu2024dare}, these methods create significant risks as the large-magnitude weights from a domain-specific model can easily overwrite the smaller, yet more critical weights for long CoT capability. 
Other works \citep{liu2025sens,Amin25AIM} utilize the product of weight magnitude and its gradient on a certain domain to identify how crucial the model weight is. 
Some do this by identifying key neurons to preserve crucial knowledge \citep{ma2025led} while others resolve knowledge conflicts before merging \citep{sun2025cat,Sun2025TATR}.
However, domain-specific gradient is not a suitable proxy for long CoT, as they often track performance adjustments on certain domains instead of the multi-step reasoning deduction that is crucial for long CoT capability.
These superficial gradients make it challenging to identify and preserve the specific weights that are essential to long CoT capability \citep{Thapa2025DisentanglingRA,hao2025training,zeng25revisiting}.
Consequently, merged models through these methods inadvertently compromise long CoT capability. Moreover, as shown in Figure \ref{fig:gibberish_rate}, these models lead to the generation of non-sensical gibberish outputs, highlighting the primary challenge of improving performance in a specific domain without sacrificing long CoT capability.

Motivated by this objective, we propose our core method: Merging Long Chain-of-Thought Models with Domain-Specific Models by Considering \textbf{R}easoning \textbf{C}apability as \textbf{P}rior (\textbf{RCP-Merging})
RCP-Merging is a framework designed to equip a domain-specific model with long CoT capability by merging with a reasoning model and further boosting the merged model's performance on its original domain-specific tasks.
The cornerstone of our method is the \textbf{Reasoning Preservation Indicator}. 
Instead of relying on conventional methods focusing on the gradient of loss on a certain domain and the magnitude of model weight, our method treats the model's long CoT capabilities as a guiding principle for the merge. 
It conceptually views reasoning model's parameters as a stable prior, constraining updates that would significantly deviate from this established reasoning foundation using the Fisher Information Matrix \citep{fisher1925theory} gained from each calibration data. 
This ensures that as the model acquires new domain-specific knowledge, it is given an indicator for each model weight to ensure the merged weight does not greatly harm the long CoT capability, consistently preventing catastrophic forgetting, gibberish output, and long CoT capability degradation that emerged from previous methods.
Our framework complements this with \textbf{Domain Knowledge Sensitivity} to identify and retain important domain-specific weights. 
Finally, \textbf{Reasoning-preserved Merging} step synthesizes these factors, utilizing both the reasoning preservation matrix and domain sensitivity as a comprehensive guide to select the most critical parameters for the final model, achieving a robust balance between domain-specific knowledge and long CoT capability.

We demonstrate that RCP-Merging, requiring only a small number of open-source calibration samples, can effectively integrate long CoT reasoning capabilities into a domain-specific model. 
Through extensive experiments across various tasks and model architectures like Qwen2.5 \citep{Yang24qwen} and Llama3.1 \citep{Aaron24llama}, our method consistently produces merged models that not only preserve domain-specific expertise but also exhibit surprisingly long CoT capabilities when addressing domain-specific questions, ultimately elevating their performance in certain domains.
Notably, the average performance of the merged model on eight datasets improves by 9.5\% and 9.2\% compared with the state-of-the-art method on BioMedicine and Finance domains, respectively. 
Moreover, though model merging aims to find a comprehensive model that compromises the performance of original models, our method improved performance by 4.5\% and 0.7\% on PubMedQA and MedQA datasets \citep{Jin19PubMedQA,Jin20MedQA}, respectively, and improved performance by 0.5\% on ConvFinQA dataset \citep{cheng2024adapting} compared to the original domain-specific models. 
To sum up, our contributions include: 

\begin{itemize}
    \item We propose a novel model merging framework, RCP-Merging, which effectively integrates a domain-specific model with a long CoT reasoning model by treating reasoning ability as a prior.
    \item We conduct extensive experiments across multiple benchmarks, demonstrating that RCP-Merging surpasses existing methods by preserving both specialized knowledge and long-CoT reasoning capabilities.
    \item Results surprisingly demonstrate that models merged via RCP-Merging exhibit emergent long CoT reasoning capabilities within model outputs when handling domain-specific problems.
\end{itemize}

%% file: related_work.tex
\section{Related Work}
Model merging \citep{goddard-etal-2024-arcees-mergekit, yang2024modelmerge,ruan2023fromtask,li2023deepmodelfusion,lu2024mergeensemblecooperate} aims to combine multiple specialized models into a single, powerful model without costly retraining \citep{ilharco2023editing,yadav2023tiesmerging,yang2023adamerge,alexandrov-etal-2024-mitigating}. Existing approaches can be broadly categorized based on the information they use to determine how parameters are combined: magnitude-based methods that operate directly on parameter values, and activation-based methods that leverage model outputs or gradients on calibration data.
\subsection{Magnitude-Based Methods}
Magnitude-based methods
merge models by performing arithmetic operations directly on their weight parameters or task vectors, often using parameter magnitude as a proxy for importance.

A foundational approach is simple Linear or weight averaging, which calculates the element-wise mean of the parameters of all models to be merged \citep{Izmailov2018Averaging,Wortsman2022ModelSoup}. 
Task Arithmetic \citep{ilharco2023editing} refines this by first computing task vectors, defined as the difference between fine-tuned and pre-trained weights ($\delta_{ft} = \theta_{ft} - \theta_{pre}$). 
These vectors, representing task-specific knowledge, are then combined through arithmetic operations like addition or negation before being applied to the base model.


To mitigate interference between task vectors, several methods have been proposed. 
TIES-Merging \citep{yadav2023tiesmerging}  introduces a three-step process: it trims each task vector by retaining only a top-k of high-magnitude parameters and resetting the rest to zero, then elects a single, dominant sign for each parameter across all task vectors. 
DARE \citep{yu2024dare} and PCB-Merging \citep{du2024pcb} adjust model weights to reduce task conflicts
by randomly dropping a ratio of weights and rescaling the remaining ones. 
FuseLLM \citep{wan24fusellm} operates by leveraging the generative probability distributions of diverse source LLMs to externalize their knowledge, which is then transferred to a single target model through a lightweight continual training phase.

A primary drawback of magnitude-based methods is their assumption that parameter magnitude equates to importance. 
This can lead to the retention of high-magnitude parameters that are harmful to other models, causing significant knowledge conflicts and degrading the performance of the merged model.

\subsection{Activation-Based Methods}
To address the limitations of magnitude-based approaches, activation-based methods leverage data-driven signals, such as model activations or gradients on a small calibration set, to obtain a more nuanced understanding of parameter importance \citep{springenberg2014striving,shrikumar2017learning,Sundararajan2017Axiomatic_Attribution_for_Deep_Networks,michel2019sixteen,maini2023can,wang-etal-2023-label,liu2024devil}.

Sens-Merging \citep{liu2025sens} operates at two levels to perform task-specific analysis to identify the sensitivity of each layer and evaluate cross-task transferability between different models on a calibration dataset. 
CAT-Merging \citep{sun2025cat} directly tackles knowledge conflict \citep{Sun2025TATR} by identifying and trimming conflict-prone components from task vectors. 
Using a few unlabeled examples, it computes layer-specific projection operators for linear weights and masks for normalization parameters to resolve interference before merging. 

Moreover, Fisher Merging \citep{matena2022fisher} and RegMean \citep{jin2023regmean} using Fisher Information Matrix to determine parameter importance or utilizing local regression for model merging; however, these approaches are characterized by high computational complexity.
Other methods, such as Activation-Informed Merging (AIM) \citep{Amin25AIM}  and LED-Merging \citep{ma2025led} utilize activations to guide the merging process, offering ways to find neurons that are crucial to certain domains. 


While these activation-based methods can more effectively mitigate the knowledge conflicts seen in magnitude-based approaches, they have their own limitations since the gradient-based evaluation is hard to capture the complex, sequential reasoning patterns within the model's weight. 

%% file: Preliminary.tex
\section{Preliminary}

\subsubsection{Task Vector.}
We adopt the concept of task vectors from the field of model merging. A task vector, $\delta$, represents the knowledge acquired by a model during fine-tuning for a specific task. 
It is computed as the difference between the weights of the fine-tuned model and base model, $\theta_{t}$, where $t$ represents the domain-specific task. 
The weights of the original pre-trained base model is represented by $\theta_{pre}$:
\begin{equation}
    \delta_t = \theta_{t} - \theta_{pre}, \ \text{for} \ t \ \in\{1,...,T\}.
    \label{eq:task_vector}
\end{equation}
In our framework, we define a task vector for each domain-specific model, $\delta_t = \theta_t - \theta_{pre}$, and a task vector for the reasoning model, $\delta_r = \theta_r - \theta_{pre}$, where $\theta_t$ and $\theta_r$ are the weights of the domain-specialized model and the long-chain reasoning model, respectively. Task vector-based merging combines these task vectors into a single, static model:
\begin{equation}
    \theta_{merged} = \theta_{pre}+\sum_{t=1}^T\lambda \cdot \delta_t,
    \label{eq:task_vector}
\end{equation}
where the coefficient $\lambda$ represents the importance of each merged task vector.

\subsubsection{Fisher Information Matrix.}
The Fisher Information Matrix (FIM) is a fundamental concept in information geometry that quantifies the amount of information an observable random variable, $x$, carries about an unknown parameter, $\theta$, of a statistical model.
For a model with parameters $\theta$, the FIM element $F(\theta)_{ij}$ is defined as the expected value of the outer product of the gradients of the log-likelihood function, the $(i, j)$-th element of the matrix can be denoted as:
\begin{equation}
    F(\theta)_{ij} = E_{x \sim p(x|\theta)} \left[ \left( \frac{\partial}{\partial\theta_i} \log p(x|\theta) \right) \left( \frac{\partial}{\partial\theta_j} \log p(x|\theta) \right) \right].
    \label{eq:fim_definition}
\end{equation}
This can also be expressed as the negative expected value of the Hessian of the log-likelihood:
\begin{equation}
    F(\theta)_{ij} = -E_{x \sim p(x|\theta)} \left[ \frac{\partial^2}{\partial\theta_i \partial\theta_j} \log p(x|\theta) \right].
    \label{eq:fim_hessian}
\end{equation}
In the context of autoregressive tasks where the loss function $\mathcal{L}(\theta, x)$ is the negative log-likelihood, i.e., $\mathcal{L}(\theta, x) = -\log p(x|\theta)$, the diagonal elements $F_i$ of the FIM can be approximated by the expected squared gradient of the loss function. 
For a single $i$-th diagonal parameter $\theta_i$ and a dataset $D_r$, this approximation is:
\begin{equation}
    F(\theta)_i \approx E_{d \sim D_r} \left[ \left( \frac{\partial \mathcal{L}(\theta, d)}{\partial \theta_i} \right)^2 \right] = E_{d \sim D_r} \left[ (g_{i,d})^2 \right],
    \label{eq:fim_approx}
\end{equation}
where $g_{i,d}^r$ is the gradient of the loss with respect to the parameter $\theta_r^i$ for a given data sample $d$. This approximation is pivotal for calculating our reasoning capability indicator.

%% file: methodology.tex
\begin{figure*}[t]
\centering
\includegraphics[width=0.90\textwidth]{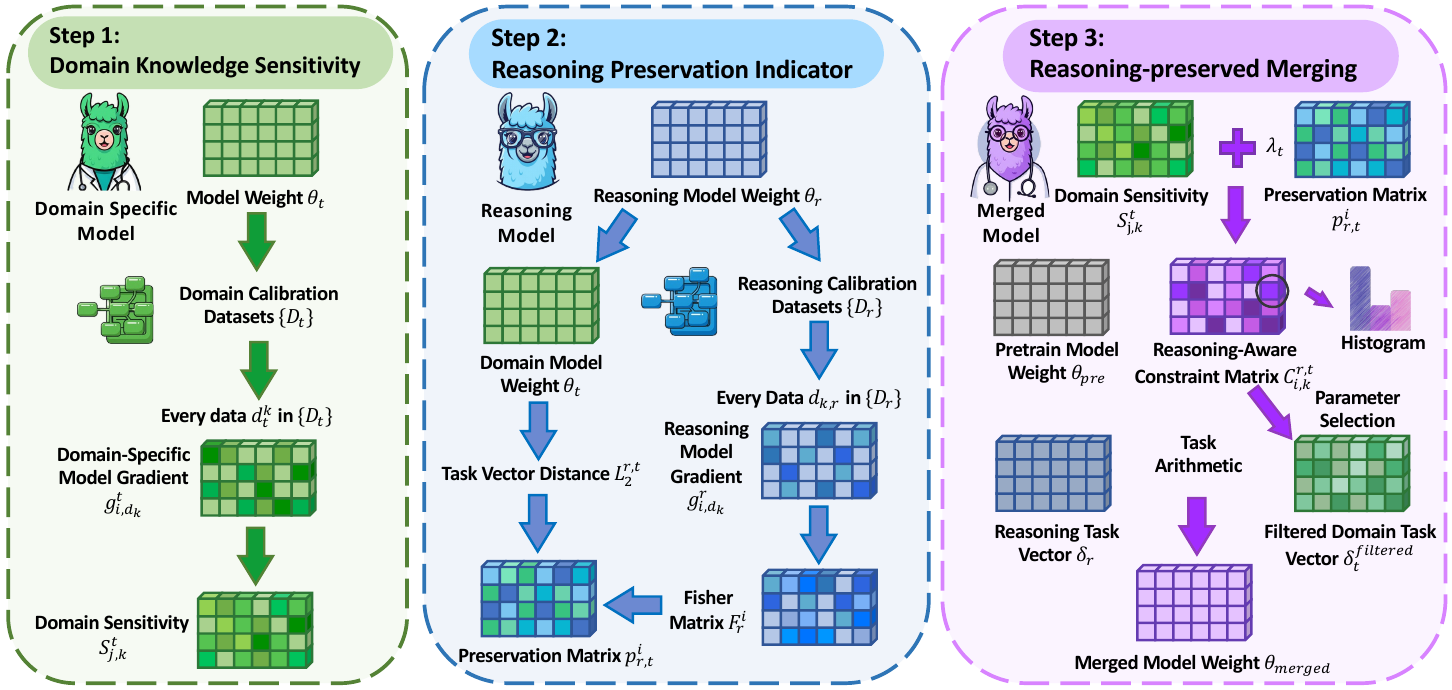}
\caption{RCP-Merging consists of three stages. (1) \textbf{Domain Knowledge Sensitivity}. This step quantifies each weight's importance for a specific domain by measuring the change in model loss when that weight is removed. (2) \textbf{Reasoning Preservation Indicator}. To protect the model's core reasoning capabilities, this stage applies a preservation term to weights that are crucial for reasoning. (3) \textbf{Reasoning-preserved Merging}. The final stage balances domain sensitivity and the reasoning preserving matrix, merging only the weights that enhance domain knowledge without harming reasoning capabilities.}
\label{fig:main_algorithm}
\end{figure*}
\section{Methodology}

Our methodology is designed to merge models by integrating domain-specific knowledge while preserving long CoT capability. 
This is achieved by first identifying parameters crucial for domain-specific tasks and then applying a preservation term derived from the Bayesian rule to mitigate the degradation of reasoning abilities. 
The final model is constructed by selectively merging domain-specific task vectors based on a reasoning-aware constraint matrix, as shown in Figure \ref{fig:main_algorithm}.

\subsection{Domain Knowledge Sensitivity}
To quantify the importance of each parameter on the domain-specific model for a task $t$, by setting the corresponding model as the domain-specific task model $\theta_t$, we introduce the concept of Domain Knowledge Sensitivity, $S_{i,k}^t$. 
This metric measures the impact on the model's performance when a particular weight is nullified.

Given a domain-specific model with parameters $\theta_t = [\theta_1, \theta_2, \dots, \theta_N]$ and a calibration dataset $\{D_t\}$, the sensitivity of the $i$-th parameter $\theta_t^i$ with respect to a data sample $d_t^k \in D_t$ is defined as the change in the loss function:
\begin{equation}
    S_{i,k}^t = \left[\mathcal{L}(\theta_t) - \mathcal{L}(\theta_t - \theta_t^i)\right]_{d=d_t^k},
    \label{eq:sensitivity_def}
\end{equation}
where $\theta_t^i$ is a vector with only the $i$-th parameter being non-zero.

For computational efficiency, we approximate this value using a first-order Taylor expansion. 
This simplifies sensitivity to the product of the parameter and its corresponding gradient, $g_{i,d_k}^t = \frac{\partial \mathcal{L}(\theta_t)}{\partial \theta_t^i}$, as follows:
\begin{equation}
\begin{aligned}
    S_{i,k}^t &\approx \left\| \frac{\partial \mathcal{L}(\theta_t)}{\partial \theta_t^i} \cdot \theta_t^i \right\|_{d=d_t^k}  
    \approx \| g_{i,d_k}^t \cdot \theta_t^i \|_{d=d_t^k}. \label{eq:sensitivity_grad,eq:sensitivity_taylor} 
\end{aligned}
\end{equation}
A lower sensitivity score indicates that the parameter $\theta_t^i$ contributes positively to the model's performance in the specific domain, as its presence reduces the loss.

\subsection{Reasoning Preservation Indicator}

To prevent the primary drawback of output collapse emerging from previous methods when merging with reasoning models, we introduce a preserving function to indicate important weights in the model merging process. 
Inspired by \citet{James16ewc}, we adopt the Bayesian rule where the reasoning model's parameter distribution serves as a prior for the posterior distribution of the final merged model's parameters. 
This approach constrains the weights to remain close to values crucial for reasoning, a detailed derivation is available in Appendix A. \ref{sec:appendix_bayes} 
Our goal is to find the parameters $\theta_t$ that maximize the posterior probability (MAP estimation), which is equivalent to minimizing the negative log-posterior: 
\begin{equation}
    \theta_{MAP} = \arg\min_{\theta_t} [-\log P(D_t|\theta_t) - \log P(\theta_t|D_r)].
    \label{eq:map_estimation}
\end{equation}
The term $-\log P(\theta_t|D_r)$ acts as a regularization term, discouraging the parameters from deviating significantly from the optimal weights learned on the reasoning task, which we denote as $\theta_r^*$.

However, directly computing the true posterior $P(\theta_t|D_r)$ is intractable for complex neural networks. 
To address this, we employ the Laplace approximation, which approximates the posterior with a Gaussian distribution centered at the mode $\theta_r^*$: $P(\theta_t|D_r) \approx \mathcal{N}(\theta_t|\theta_{r}^*,F_{r}^{-1})$. 
The precision matrix of this Gaussian is the Fisher Information Matrix (FIM), $F_r$, which measures the curvature of the log-likelihood landscape. The probability density function is:
\begin{equation}
P(\theta_t|D_r) \approx \frac{|F_r|^{1/2}}{(2\pi)^{k/2}} \exp\left(-\frac{1}{2}(\theta_t - \theta_r^*)^T F_r (\theta_t - \theta_r^*)\right)
\label{eq:probability_density_function}
\end{equation}
By taking the natural logarithm and discarding terms that are constant with respect to $\theta_t$, we simplify the expression for optimization purposes. This yields a tractable form for the log-posterior preservation matrix:
\begin{equation}
    \log P(\theta_t|D_r) \approx -\frac{1}{2}(\theta_t - \theta_r^*)^T F_r (\theta_t - \theta_r^*).
    \label{eq:log_posterior}
\end{equation}
This quadratic term measures how much the updated parameters $\theta_t$ have diverged from the reasoning-optimal parameters $\theta_r^*$, weighted by the FIM $F_r$. A higher value in $F_r$ for a certain parameter indicates its importance for the reasoning task,
and thus incurs a larger preservation for any deviation.


To make this computation more feasible, we assume a diagonal FIM. 
As shown in Equation \ref{eq:fim_approx}, this simplifies $p^i_{r,t}$ into a sum of per-parameter contributions, where for each parameter $\theta_t^i$, the penalty is $\log P(\theta_t^i|D_{r}) \approx -\frac{1}{2}F_{r,ii}(\theta_t^i - \theta_{r,i}^*)^2$. 
The $i$-th diagonal elements of the FIM, $F_{r,ii}$, can be approximated by the average of the squared gradients over the calibration reasoning dataset $D_r = \{d_k\}_{k=1}^{N_r}$. 
Combining these steps, we define the final reasoning preservation indicating matrix $p^i_{r,t}$ for each parameter $\theta^i$ as:
\begin{equation}
    p^i_{r,t} = -\left\| \frac{1}{2N_r} \sum_{k=1}^{N_r} (g_{i,d_k}^r)^2 (\theta_t^i - \theta_r^i)^2 \right\|.
    \label{eq:penalty_final}
\end{equation}
Here, $g_{i,d_k}^r$ is the gradient of the loss for sample $d_k$ with respect to parameter $\theta_r^i$. This metrics quantifies how much the new parameter $\theta_t^i$ impairs the model's reasoning ability.

\subsection{Reasoning-preserved Merging}

To integrate domain knowledge while preserving core reasoning skills, we propose reasoning-aware merging strategy. 
We implement this by defining a Constraint metric $C_{i,k}^{r,t}$ for each parameter $\theta_i$ to quantify the importance of long CoT capability, combining its Domain Knowledge Sensitivity ($S_{i,k}^t$) and Reasoning Capability Indicator ($p^i_{r,t}$):
\begin{equation}
    C_{i,k}^{r,t} = S_{i,k}^t + \lambda_r \cdot p^i_{r,t}.
    \label{eq:constraint_metric}
\end{equation}
Here, the hyperparameter $\lambda_r$ balances the trade-off between domain performance and long CoT capability preservation.

Next, we filter parameter updates using a majority vote criterion. 
An update for parameter $\theta_t^i$ is accepted if more data samples in the domain dataset $D_t$ yield a negative conflict score than a non-negative one:
\begin{equation}
    \text{Accept update for } \theta_t^i \quad \text{if} \quad N(C_{i,k}^{r,t} < 0) > N(C_{i,k}^{r,t} \geq 0).
    \label{eq:filter_condition}
\end{equation}
This condition generates a binary mask $M \in \{0, 1\}^N$, where $M_i=1$ signifies an accepted update for the corresponding parameter.

Finally, we use this mask to create a filtered domain-specific task vector, $\delta_t^{filtered}$, via an element-wise product with the original task vector $\delta_t = \theta_t - \theta_{pre}$. The final model weights, $\theta_{merged}$, are then obtained by adding the complete reasoning vector $\delta_r$ and the weighted sum of these filtered task vectors to the pre-trained weights $\theta_{pre}$:
\begin{align}
    \delta_t^{filtered} &= M \odot \delta_t, \label{eq:filtered_vector} \\
    \theta_{merged} &= \theta_{pre} + \delta_r + \sum_{t=1}^{T} \lambda_t \cdot \delta_t^{filtered}, \label{eq:final_merge} 
\end{align}
where $T$ is the number of domain-specific tasks and $\lambda_t$ are scaling coefficients. This approach ensures the model benefits from domain-specific knowledge while robustly maintaining its reasoning abilities.


%% file: experiment.tex
\section{Experiment}

\input{table/main_exp}

\subsection{Experimental Setup}

\subsubsection{Baselines.}

We compare RCP-Merging with multiple merging baselines: \textbf{Average} \citep{Izmailov2018Averaging}, \textbf{Task Arithmetic} \citep{ilharco2023editing}, \textbf{TIES-Merging} \citep{yadav2023tiesmerging}, \textbf{DARE-Merging}, \textbf{DARE-Merging} with TIES \citep{yu2024dare}, \textbf{FuseLLM} \citep{wan24fusellm}, \textbf{FuseLLM with AIM}, \textbf{DARE TIES with AIM} \citep{Amin25AIM}, \textbf{Sens-Merging} \citep{liu2025sens}, and \textbf{CAT-Merging} \citep{sun2025cat}. 
We utilize mergekit \citep{goddard-etal-2024-mergekit} as merging tools for baseline methods, detailed discussions and recommended hyperparameters are listed in Appendix B.1 \ref{app:baselines} and B.2. \ref{app:hyper_setting} 


\subsubsection{Datasets\&Metrics.}
We assess merged model performance through four pillars: 
(1) Mathematical reasoning (Math) via GSM8k \citep{Cobbe21gsm} and AIME2024 \citep{aime_1983_2024} (Accuracy$\uparrow$ with CoT); 
(2) Code generation (Code) evaluated by HumanEval \citep{Mark21humaneval} and LiveCodeBench \citep{jain2024livecodebench} (Pass@1$\uparrow$); 
(3) Medical question answering (BioMedicine) through PubMedQA \citep{Jin19PubMedQA} and MedQA \citep{Jin20MedQA} (Accuracy$\uparrow$);  
(4) General knowledge question answering with ARC-C \citep{Clark18arc} and GPQA \citep{rein2023gpqa} (Accuracy$\uparrow$). 

\subsubsection{Models.}
The experiment involves a set of models built upon the Qwen2.5-7B  \citep{Yang24qwen} Base model architecture. 
The domain-specific model is Meditron3-Qwen2.5-7B \citep{Chen23Meditron} for BioMedicine, and the Reasoning model is DeepSeek-R1-Distill-Qwen-7B \citep{guo2025deepseek}.

\subsection{RCP-Merging's Superior Performance}

RCP-Merging achieves SOTA average performance on BioMedicine domain, surpassing all existing merging methods and even the original BioMedicine model meanwhile maintaining long CoT capability.
The results, summarized in Table \ref{tab:performance_comparison}, demonstrate that RCP-Merging achieves a superior balance between domain-specific expertise and reasoning capabilities. 
It obtains the highest average score of 49.4 across all benchmarks, significantly surpassing all other merging methods. Specifically, in the target BioMedicine domain, RCP-Merging achieves top scores on both PubMedQA and MedQA with scores of 55.5 and 54.1, effectively integrating BioMedicine knowledge. 

Simultaneously, it not only preserves but also enhances the reasoning abilities inherited from the reasoning model, leading to state-of-the-art performance among the merged models in Math with 84.3 on GSM8K, 33.3 on AIME2024 and Code shown as the performance of 71.3 on HumanEval. 
In contrast, most baseline methods struggle to maintain this balance, sacrificing reasoning for domain performance. 
This highlights RCP-Merging's unique effectiveness in creating a truly versatile and capable model. 
We also utilizing the merged model's output content length to represent the reasoning thinking process of the merged models, the detailed analysis and results are shown in Table \ref{tab:generated_text_length} in Appendix.
To further analysis the merged model's reasoning process, we also conduct case study in Appendix. \ref{app:case_study}


\begin{figure*}[h]
\centering
\includegraphics[width=0.95\textwidth]{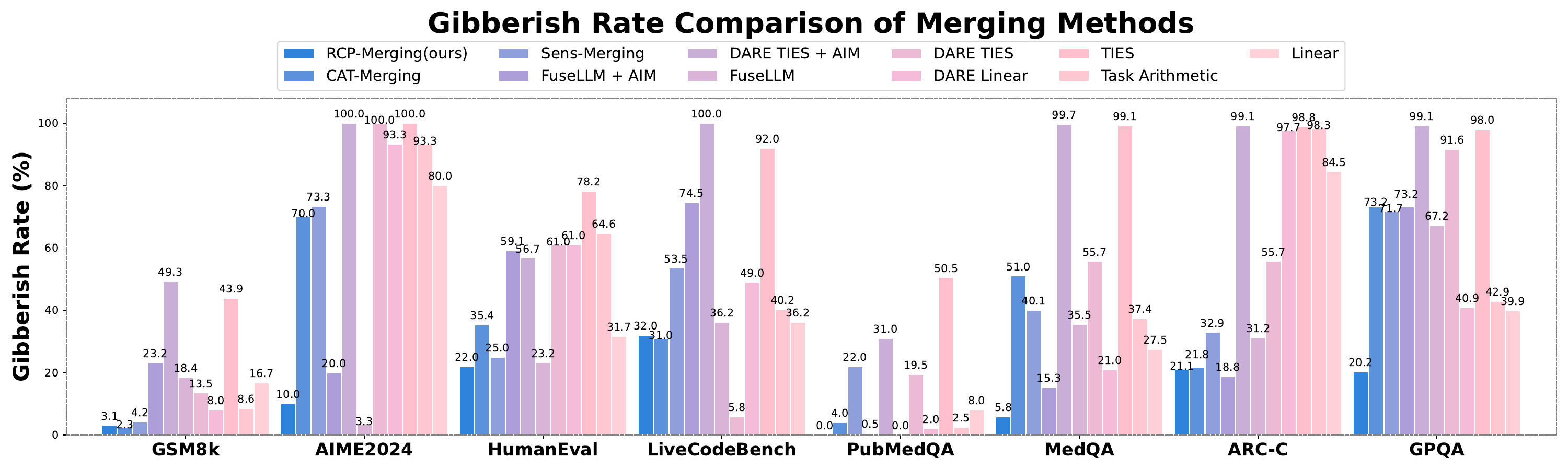}
\caption{
Gibberish rate comparison for merging Qwen2.5-7B (Base), Meditron3-Qwen2.5-7B (BioMedicine), and DeepSeek-R1-Distill-Qwen-7B (Reasoning) on all datasets, where a lower rate indicates higher-quality content.
}
\label{fig:gibberish_rate}
\end{figure*}

\subsubsection{RCP-Merging's Output Stability.}


To address model stability, we measure gibberish rate: the frequency of nonsensical outputs identified by a GPT\-4 evaluator \citep{openai2023gpt4} to validate genuine performance against output degeneration. 
As shown in Figure \ref{fig:gibberish_rate}, RCP-Merging demonstrates superior stability, achieving a low 14.3\% average gibberish rate (0\% on PubMedQA, 5.8\% on MedQA). This starkly contrasts with baseline methods like TIES (82.3\%) and DARE TIES  AIM (79.5\%), which suffer from significant output collapse. This confirms RCP-Merging's robust performance stems from genuine capability integration. 

\subsection{Different Domain-specific Task}
To verify the generalizability of our method across different domains, we conduct experiments where specific domain is shifted from BioMedicine to finance. 
In this setup, we merge WiroAI-Finance-Qwen-7B as Finance model
with the same Base and Reasoning models.

RCP-Merging demonstrates top performance across all evaluated categories, including GSM8k in Math, HumanEval in Code, ARC-C in Knowledge, and ConvFinQA in Finance \citep{cheng2024adapting}.
Shown in Table \ref{tab:finance_performance}, results demonstrate that RCP-Merging achieves the highest average score of 72.2, decisively outperforming all baseline methods on Finance domain. 
As the performance shown in the table, RCP-Merging demonstrates the best performance across four benchmarks, this further verifies the scalability of RCP-Merging in different fields.
Results show RCP-Merging balances domain-specific performance and long CoT capability across multiple domains.
\input{table/finance}

\subsection{Different Model Architecture.}

RCP-Merging demonstrates consistent performance across different architectures. 
We have verified this by conducting experiments on the Llama3.1-8B based models, which is distinct from our primary setup.
In this alternative configuration, we used Llama3.1-8B \citep{Aaron24llama} as the Base model, Llama3-OpenBioLLM-8B \citep{OpenBioLLMs} as the BioMedicine model, and DeepSeek-R1-Distill-Llama-8B \citep{guo2025deepseek} as the Reasoning model.

We use GSM8k, HumanEval, ARC-C and PubMedQA to indicate the performance of different merge methods on Math, Code, Knowledge and BioMedicine domain.
As the results in Table \ref{tab:llama_8b_performance}, RCP-Merging achieves the best average score of 68.3 among all merging techniques. 
Although FuseLLM with AIM shows a slightly better score in the specific BioMedicine domain, RCP-Merging has the best overall capability. 
\input{table/llama}


%% file: table/main_exp.tex
\begin{table*}[!t]
	\centering
	{
	\setlength{\tabcolsep}{2pt}
	
	\begin{tabular}{l|cc|cc|cc|cc|c}
		\toprule
		\multirow{2}{*}{\textbf{Method/Task}} & \multicolumn{2}{c|}{\textbf{Math}} & \multicolumn{2}{c|}{\textbf{Code}} & \multicolumn{2}{c|}{\textbf{BioMedicine}} & \multicolumn{2}{c|}{\textbf{Knowledge}} & \multirow{2}{*}{\textbf{Average}} \\
		\cmidrule(lr){2-3} \cmidrule(lr){4-5} \cmidrule(lr){6-7} \cmidrule(lr){8-9}
		& \textbf{GSM8K} & \textbf{AIME2024} & \textbf{HumanEval} & \textbf{LiveCodeBench} & \textbf{PubMedQA} & \textbf{MedQA} & \textbf{ARC-C} & \textbf{GPQA} & \\
		\midrule
		\textbf{Base}        & 69.4 & 0.0  & 50.6 & 12.4 & 32.5 & 22.9 & 60.9 & 7.6  & 32.0 \\
		\textbf{BioMedicine} & 81.5 & 0.0  & 54.3 & 2.2  & 51.0 & 53.5 & 74.9 & 9.6  & 40.9 \\
		\textbf{Reasoning}   & 86.7 & 56.7 & 76.6 & 29.8 & 38.0 & 30.2 & 76.5 & 15.2 & 51.2 \\
		\midrule
		Linear             & 46.4 & 0.0  & 32.3 & 2.8  & 34.0 & 20.2 & 32.6 & \textbf{15.7} & 23.0 \\
		Task Arithmetic    & 63.5 & 0.0  & 21.3 & 2.8  & 31.0 & 39.3 & 27.7 & 14.7 & 25.0 \\
		TIES-Merging       & 40.6 & 0.0  & 39.6 & 1.3  & 22.5 & 22.5 & 25.8 & 1.0  & 19.2 \\
		DARE Linear        & 70.2 & 0.0  & 49.4 & 2.5  & 31.5 & 37.3 & 29.5 & 14.7 & 29.4 \\
		DARE TIES          & 43.1 & 0.0  & 38.4 & 4.3  & 24.5 & 41.4 & 24.6 & 8.6  & 23.1 \\
		FuseLLM            & 41.5 & 26.7 & 53.7 & 5.0  & 35.0 & 27.3 & 57.2 & 9.6  & 32.0 \\
		DARE TIES \& AIM   & 37.8 & 0.0  & 18.9 & 0.4  & 22.0 & 19.3 & 24.2 & 2.0  & 15.6 \\
		FuseLLM \& AIM     & 40.3 & 20.0 & 26.2 & 3.8  & 20.5 & 25.5 & 57.8 & 7.6  & 25.2 \\  
		Sens-Merging       & 79.8 & 16.7 & 53.0 & \textbf{21.7} & 34.0 & 46.6 & 59.4 & 8.1  & 39.9 \\
		CAT-Merging        & 60.7 & 10.0 & 39.0 & 20.1 & 39.0 & 40.4 & 60.7 & 8.1  & 34.8 \\
		\rowcolor{lightgray!25} \textbf{RCP-Merging} & \textbf{84.3} & \textbf{33.3} & \textbf{71.3} & 18.4 & \textbf{55.5} & \textbf{54.1} & \textbf{82.5} & \textbf{15.7} & \textbf{49.4} \\
		\bottomrule
	\end{tabular}
	} 
	\caption{Performance comparison of merging Qwen2.5-7B (Base), Meditron3-Qwen2.5-7B (BioMedicine) and DeepSeek-R1-Distill-Qwen-7B (Reasoning) on all datasets across Math, Code, BioMedicine, and Knowledge areas. The best performance among all merging methods on each dataset is highlighted in \textbf{bold}.}
	\label{tab:performance_comparison}
\end{table*}

%% file: table/finance.tex
\begin{table}[H]
	\centering

    \resizebox{0.98\columnwidth}{!}{
	\begin{tabular}{l|cccc|c}
		\toprule
		\textbf{Method/Task} & \textbf{Math} & \textbf{Code} & \textbf{Finance} & \textbf{Knowledge} & \textbf{Average} \\
		\midrule

		\textbf{Base}            & 69.4 & 50.6 & 50.3 & 60.9 & 57.8 \\
		\textbf{Finance}         & 50.2 & 1.2  & 58.7 & 47.9 & 39.5 \\
		\textbf{Reasoning}       & 86.7 &    76.8 & 36.2 & 76.5 & 69.1 \\
		\midrule
		Linear          & 16.6 & 32.3 & 34.0 & 27.7 & 27.7 \\
		Task Arithmetic & 8.4  & 39.6 & 17.4 & 43.3 & 27.2 \\
		TIES-Merging            & 7.2  & 21.3 & 18.8 & 42.0 & 22.3 \\
		DARE Linear     & 8.4  & 49.4 & 17.7 & 43.1 & 29.7 \\
		DARE TIES       & 7.6  & 38.4 & 18.4 & 43.6 & 27.0 \\
		FuseLLM         & 7.4  & 53.7 & 18.4 & 42.7 & 30.6 \\
		DARE TIES \& AIM   & 6.4  & 18.9 & 19.7 & 46.5 & 22.9 \\
		FuseLLM \& AIM    & 5.3  & 26.2 & 20.4 & 47.1 & 24.8 \\
		Sens-Merging      & 60.7 & 53.7 & 4.2  & 25.8 & 36.1 \\
		CAT-Merging       & 60.7 & 39.0 & 10.1 & 24.8 & 33.7 \\
		\rowcolor{lightgray!25} 
		\textbf{RCP-Merging}  & \textbf{82.0} & \textbf{71.3} & \textbf{59.2} & \textbf{76.4} & \textbf{72.2} \\
		\bottomrule
	\end{tabular}}
    	\caption{Performance comparison of merging Qwen2.5-7B (Base), WiroAI-Finance-Qwen-7B  (Finance) and DeepSeek-R1-Distill-Qwen-7B (Reasoning) on four datasets across Math, Code, Finance and Knowledge areas. The best performance among all merging methods on each dataset is highlighted in \textbf{bold}.}
        \label{tab:finance_performance}
\end{table}

%% file: table/llama.tex
\begin{table}[H]
	\centering

    \resizebox{0.98\columnwidth}{!}{
	\begin{tabular}{l|cccc|c}
		\toprule
		\textbf{Method/Task} & \textbf{Math} & \textbf{Code} & \textbf{BioMedicine} & \textbf{Knowledge} & \textbf{Average} \\

		\midrule

		\textbf{Base}                & 60.9 & 42.7 & 55.0 & 60.7 & 54.8 \\
		\textbf{BioMedicine}             & 39.4 & 37.8 & 58.0 & 56.0 & 47.8 \\
		\textbf{Reasoning}           & 68.8 & 89.6 & 51.5 & 84.0 & 73.5 \\
		\midrule
		Linear          & 3.2  & 37.2 & 31.0 & 59.0 & 32.6 \\
		Task Arithmetic & 55.3 & 48.2 & 23.0 & 45.9 & 43.1 \\
		TIES-Merging            & 47.5 & 40.2 & 53.5 & 62.2 & 50.9 \\
		DARE Linear     & 58.3 & 40.2 & 23.0 & 45.9 & 41.9 \\
		DARE TIES       & 45.6 & 47.6 & 32.5 & 22.2 & 37.0 \\
		FuseLLM         & 48.8 & 61.0 & 55.5 & 53.3 & 54.7 \\
		DARE TIES \& AIM    & 38.1 & 49.4 & 13.0 & 26.0 & 31.6 \\
		FuseLLM \& AIM     & 56.1 & 59.8 & \textbf{57.5} & 59.3 & 58.2 \\
		Sens-Merging      & 65.7 & 46.3 & 55.5 & 65.5 & 58.3 \\
		CAT-Merging       & 62.5 & 55.5 & 54.0 & 64.3 & 59.1 \\
		\rowcolor{lightgray!25} \textbf{RCP-Merging} & \textbf{67.2} & \textbf{73.2} & 57.0 & \textbf{75.8} & \textbf{68.3} \\
		\bottomrule
	\end{tabular}}
        \caption{Performance comparison of merging Llama-3.1-8B (Base), Llama3-OpenBioLLM-8B (BioMedicine) and DeepSeek-R1-Distill-Llama-8B (Reasoning) on four datasets across Math, Code, BioMedicine, and Knowledge areas. The best performance under certain dataset is highlighted in \textbf{bold}.}
        \label{tab:llama_8b_performance}
\end{table}

%% file: appendix/hyperparameters.tex
\label{app:hyper}


\begin{figure}[!htbp]
  \centering
  
\raggedleft
    \begin{minipage}{0.90\columnwidth}
    
    
    \includegraphics[width=0.95\textwidth]{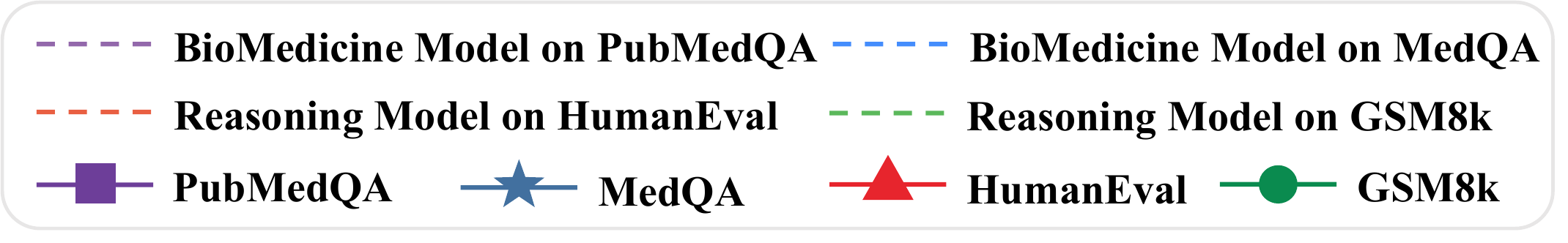}
    \label{fig:hyper_legend}
  \end{minipage}
  \vskip\baselineskip 
  \begin{minipage}{0.22\textwidth}
    \centering
    \subfigure[\centering BioMedicine Benchmark] {

     \label{fig:hyper_med}     
    \includegraphics[width=1\columnwidth]{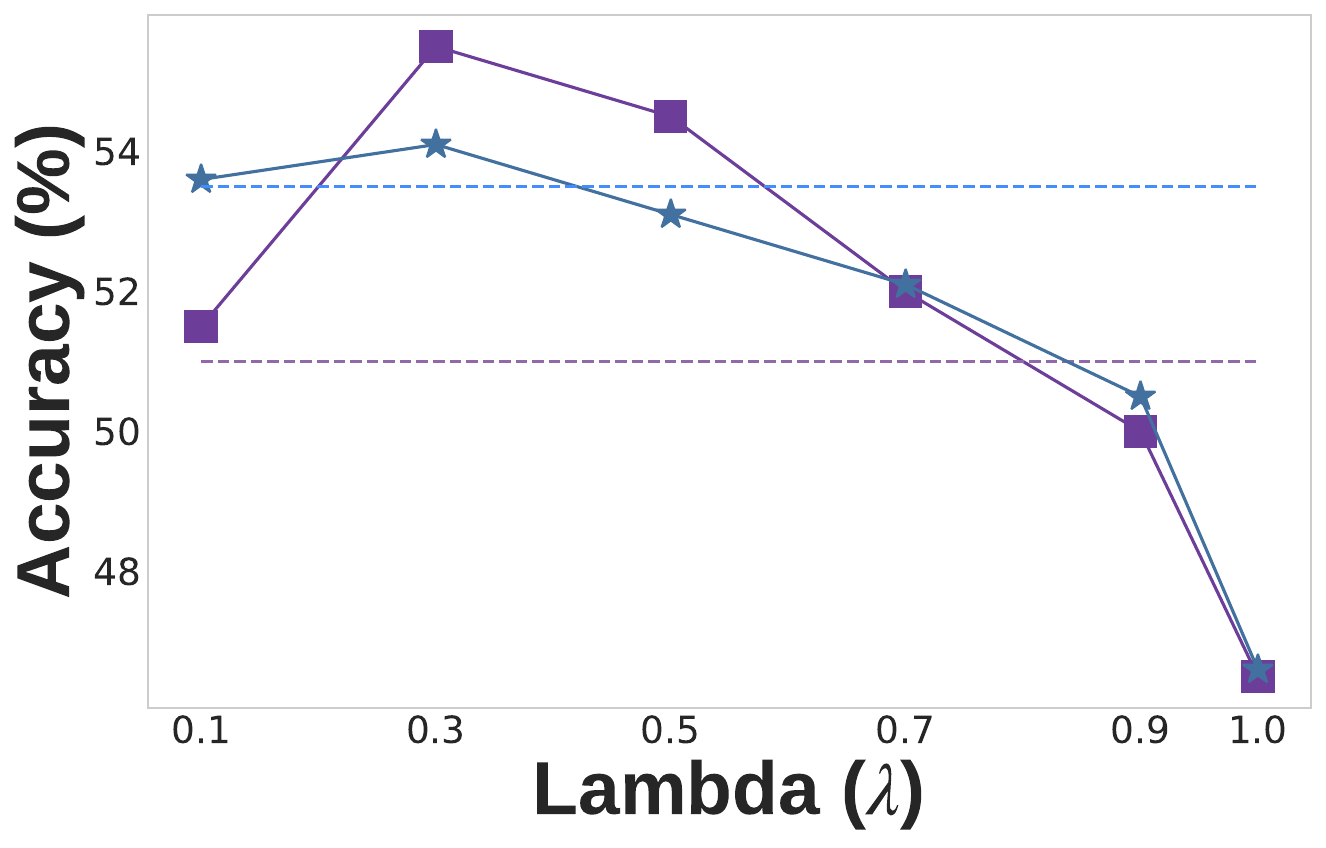}  
    }    
  \end{minipage}%
  \hfill
  \begin{minipage}{0.22\textwidth}
    \centering
    \subfigure[Reasoning Benchmark] {
     \label{fig:hyper_reason}     
    \includegraphics[width=1\columnwidth]{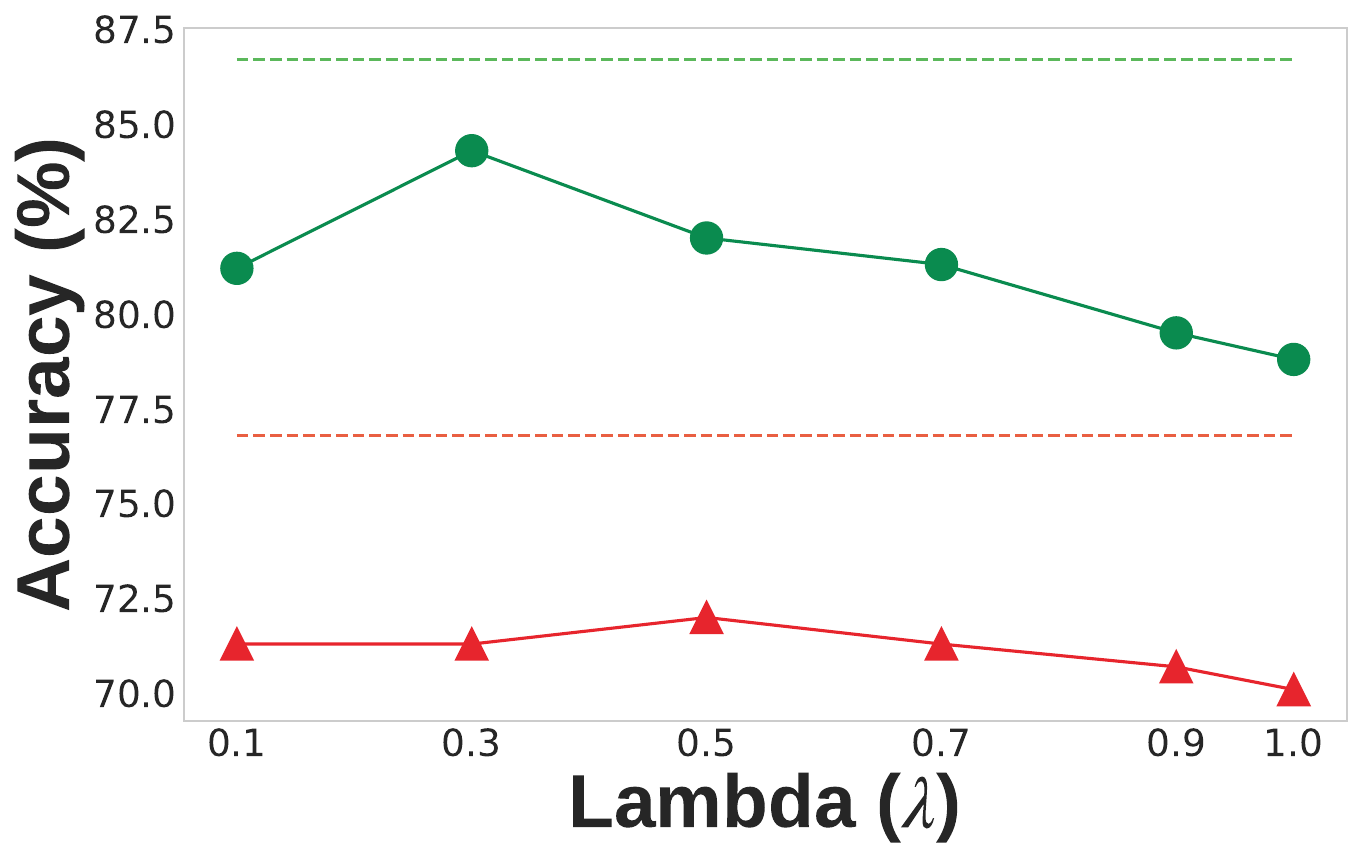}  
    }  
  \end{minipage}
  
  \caption{Hyperparameter Analysis. periments are conducted when merging Qwen2.5-7B (Base), Meditron3-Qwen2.5-7B (BioMedicine) and DeepSeek-R1-Distill-Qwen-7B (Reasoning) on BioMedicine datasets in Figure \ref{fig:hyper_med} and Reasoning datasets in Figure \ref{fig:hyper_reason}. Merged Model performance is evaluated under different Reasoning-preserving coefficients $\lambda$.}
  \label{fig:hyper_all}
\end{figure}

%% file: ablation.tex
\subsection{Ablation Study.}

This section performs an ablation study to evaluate the effectiveness of the parameter-specific trimming techniques in RCP-Merging, including the pruning of Knowledge Sensitivity and Reasoning Preservation. 
As results shown in Table \ref{tab:ablation}, excluding Domain Sensitivity ({w/o Domain Sensitivity}) causes the average score to drop significantly from 68.3 to 48.7.
The effect is even more severe when removing the Reasoning Preservation ({w/o Reasoning Preservation}), which plunges the average score to 41.4. 
These results underscore that both trimming strategies are indispensable. 

\begin{table}[H]
	\centering

    \resizebox{0.98\columnwidth}{!}{
	\begin{tabular}{l|cccc|c}
		\toprule
		\textbf{Method/Task} & \textbf{Math} & \textbf{Code} & \textbf{BioMedicine} & \textbf{Knowledge} & \textbf{Average} \\

		\midrule

		\textbf{Base}                & 60.9 & 42.7 & 55.0 & 60.7 & 54.8 \\
		\textbf{BioMedicine}             & 39.4 & 37.8 & 58.0 & 56.0 & 47.8 \\
		\textbf{Reasoning}           & 68.8 & 89.6 & 51.5 & 84.0 & 73.5 \\
		\midrule

		w/o Domain Sensitivity    &58.4		&56.1	&33.0	&47.4	&48.7 \\
        w/o Reasoning  Preservation     &  57.1	&	37.2		&30.5	&	40.9 &41.4  \\
		\rowcolor{lightgray!25} \textbf{RCP-Merging} & \textbf{67.2} & \textbf{73.2} & \textbf{57.0} & \textbf{75.8} & \textbf{68.3} \\
		\bottomrule
	\end{tabular}}
        \caption{Ablation Study. Performance comparison when merging Qwen2.5-7B (Base), Meditron3-Qwen2.5-7B (BioMedicine) and DeepSeek-R1-Distill-Qwen-7B (Reasoning) on four datasets across Math, Code, BioMedicine, and Knowledge areas. Performance The best performance under certain dataset is highlighted in \textbf{bold}.}
        \label{tab:ablation}
\end{table}

%% file: conclusion.tex
\section{Conclusion}
We propose a novel model merging framework, RCP-Merging, which effectively integrates domain-specific models with long-chain-of-thought reasoning models by treating reasoning ability as a prior. 
Our method applies a reasoning capability penalty to preserve core reasoning parameters while selectively merging essential domain-specific weights. 
Notably, RCP-Merging enhances performance in the BioMedicine and Finance domains by 9.5\% and 9.2\% respectively, compared to state-of-the-art methods.
Our approach creates powerful, unified models that excel in both domain-specific knowledge and general long-chain-of-thought reasoning, effectively addressing the challenge of balancing domain performance with reasoning capability.

%% file: appendix.tex
\newpage
\newpage
\clearpage
\appendix
\section{Appendix}

\subsection{A: Bayesian Framework}
\input{appendix/Bayesian}

\subsection{B: Experiment Details}

\subsection{B.1 Model Merging Baselines}
\input{appendix/baselinemethods}

\subsection{B.2 Details of Hyperparameters' Setting for Baselines Methods}
\input{appendix/hyperparameters_setting}

\subsection{B.3 Datasets and Evaluation Metrics}
\input{appendix/datasets}

\subsection{C: RCP-Merging's Hypterparameter Analysis}

\input{appendix/hyperparameters}
\subsection{D: RCP-Merging's Output Content Analysis}
\input{appendix/output_analysis}

\subsection{E: RCP-Merging's Output Case Study}
\input{appendix/casestudy}

\subsection{F: RCP-Merging's Generated Content Length}
\input{appendix/generated_length}

\subsection{G: Performance comparison on different Model Size}
\input{appendix/model_size}

%% file: appendix/Bayesian.tex
\label{sec:appendix_bayes}

This appendix provides a detailed derivation of the Bayesian framework used to formulate the Reasoning Capability Penalty. The central goal is to find an optimal set of parameters for a domain-specific model, denoted as $\theta_t$, by leveraging information from two sources: the new domain-specific dataset, $D_t$, and the pre-existing knowledge from a model trained on a reasoning dataset, $D_r$.

Our objective is to estimate the posterior probability distribution $P(\theta_t | D_t, D_r)$. This term represents the probability of the parameters $\theta_t$ being optimal after we have observed both the domain data $D_t$ and the reasoning data $D_r$. Maximizing this posterior probability allows us to find the most plausible parameter values.

The derivation begins with the product rule of probability, which allows us to express the joint probability $P(\theta_t, D_t | D_r)$ in two equivalent ways. First, by factoring out $\theta_t$:
\begin{equation}
    P(\theta_t, D_t | D_r) = P(D_t | \theta_t, D_r) P(\theta_t | D_r).
    \label{eq:appendix_bayes_1}
\end{equation}
Alternatively, by factoring out $D_t$:
\begin{equation}
    P(\theta_t, D_t | D_r) = P(\theta_t | D_t, D_r) P(D_t | D_r).
    \label{eq:appendix_bayes_2}
\end{equation}
By equating these two expressions, we can solve for our target posterior distribution, $P(\theta_t | D_t, D_r)$:
\begin{equation}
    P(\theta_t | D_t, D_r) = \frac{P(D_t | \theta_t, D_r) P(\theta_t | D_r)}{P(D_t | D_r)}.
    \label{eq:appendix_bayes_full}
\end{equation}
In this formulation, the term $P(\theta_t | D_r)$ plays the crucial role of the prior distribution. It encapsulates our prior belief about the parameters $\theta_t$ before encountering the new domain data $D_t$. In our method, we define this prior as the posterior distribution of the parameters obtained after training on the reasoning dataset $D_r$. Thus, its function is to act as a regularizer, ensuring that the final parameters do not stray far from the values established as important for reasoning.

The term $P(D_t | \theta_t, D_r)$ is the likelihood, which measures how probable the new domain data $D_t$ is for a given set of parameters $\theta_t$. We apply a standard conditional independence assumption, stating that the generation of new data $D_t$ depends only on the parameters $\theta_t$, not on the old data $D_r$. This simplifies the likelihood to $P(D_t | \theta_t)$.

Finally, the denominator $P(D_t | D_r)$ is the marginal likelihood or evidence. It serves as a normalization constant to ensure the posterior is a valid probability distribution. Since it does not depend on the parameters $\theta_t$ that we are optimizing, it can be disregarded when our goal is to maximize the posterior.

Considering these points, for the purpose of optimization, the posterior probability is proportional to the product of the likelihood and the prior. This leads to the final relationship used in our methodology:
\begin{equation}
    P(\theta_t|D_{t},D_r) \propto P(D_{t}|\theta_t) P(\theta_t|D_{r}).
    \label{eq:appendix_bayes_proportional}
\end{equation}
This proportionality forms the theoretical foundation for MAP (Maximum A Posteriori) estimation:
\begin{equation}
    \theta_{MAP} = \arg\max_{\theta_t} [\log P(D_t|\theta_t) + \log P(\theta_t|D_r)],
    \label{eq:map_estimation_max}
\end{equation}
where maximizing the posterior is equivalent to minimizing the negative of its logarithm, as shown in Equation \ref{eq:map_estimation}. 

%% file: appendix/baselinemethods.tex
\label{app:baselines}
This section provides detailed descriptions of the model merging baselines used in our experiments. Each method is briefly explained, highlighting its core idea and relevant formulation.

\begin{itemize}
    \item \textbf{Linear Averaging} \citep{Izmailov2018Averaging}: This basic method merges models by directly averaging their corresponding parameters. 
    \item \textbf{Task Arithmetic} \citep{ilharco2023editing}: Task Arithmetic combines task-specific knowledge by adding or subtracting parameter vectors. 
    It computes a task vector as the difference between a fine-tuned model and its base model, then scales and adds this vector to another model.
    \item \textbf{TIES-Merging} \citep{yadav2023tiesmerging}: TIES-Merging addresses parameter redundancy by identifying and merging significant parameters. 
    It involves pruning, re-scaling, and merging parameter differences to combat interference.
    \item \textbf{DARE-Merging} \citep{yu2024dare}: DARE (Drop and Restore) aims to mitigate catastrophic forgetting during merging by selectively dropping and then restoring parameters. It introduces a dropout mechanism on the parameter differences before merging.
    \item \textbf{FuseLLM} \citep{wan24fusellm}: FuseLLM proposes a method to merge large language models by aligning their activation spaces. It focuses on combining representations learned by different models rather than directly manipulating parameters.
    \item \textbf{AIM} \citep{Amin25AIM}: AIM (Activation Informed Merging) provides a more sophisticated way to align and combine the activation patterns of different models for improved merged performance.
    \item \textbf{Sens-Merging} \citep{liu2025sens}: Sens-Merging focuses on the sensitivity of model parameters to specific tasks. It aims to merge models by prioritizing parameters that are most sensitive and crucial for performance on target tasks.
    \item \textbf{CAT-Merging} \citep{sun2025cat}: CAT-Merging (Context-Aware Transformation Merging) proposes a method that considers the contextual information during the merging process. It uses a transformation function to align and combine model parameters based on their relevance to different contexts.
\end{itemize}

%% file: appendix/hyperparameters_setting.tex
\label{app:hyper_setting}
For the baseline methods, we use the following hyperparameters. In Task Arithmetic, the scaling factor is set to $\lambda=0.3$. For both TIES-Merging and DARE, the merging weight is $\lambda=0.3$ and the dropout rate is $r=0.9$. For CAT-Merging, we use $\lambda=1.0$ and $c=3$. For RCP-Merging, we use $\lambda=0.3$ as the default reasoning-preserving coefficient. During inference, we set `max\_new\_tokens' to 2048 and `temperature` to 0 for the base and task models. For the reasoning model, we use `max\_new\_tokens' of 32768, `temperature' of 0.6, and `top-k` of 0.95 for long CoT generation. 

%% file: appendix/datasets.tex
\label{app:datasets}
Our experiments evaluate merged model performance across a diverse set of datasets, categorized into four pillars to assess different capabilities. For mathematical reasoning, we use \textbf{GSM8k} \citep{Cobbe21gsm}, a dataset of grade school math word problems requiring multi-step reasoning, and \textbf{AIME2024} \citep{aime_1983_2024}, which presents advanced mathematical problems from the American Invitational Mathematics Examination, both evaluating accuracy with Chain-of-Thought (CoT). For code generation, \textbf{HumanEval} \citep{Mark21humaneval} provides programming problems that test functional correctness, while \textbf{LiveCodeBench} \citep{jain2024livecodebench} offers a dynamic and up-to-date benchmark for code generation, both measured by Pass@1. In medical question answering, \textbf{PubMedQA} \citep{Jin19PubMedQA} focuses on biomedical research questions, and \textbf{MedQA} \citep{Jin20MedQA} contains medical exam questions, with accuracy as the metric. Finally, for general knowledge question answering, \textbf{ARC-C} \citep{Clark18arc} challenges models with science questions requiring common sense reasoning, and \textbf{GPQA} \citep{rein2023gpqa} features difficult, expert-level general knowledge questions, both assessed by accuracy. These datasets collectively provide a comprehensive evaluation of the merged models' capabilities across various domains.

%% file: appendix/output_analysis.tex
\label{app:output_analysis}
To assess the linguistic quality and coherence of the generated content, we evaluate the models using two key metrics: Distinct-N \citep{li2015diversity}, which measures text diversity, and Perplexity (PPL) \citep{hu2024ppl}, which evaluates fluency. An ideal model should produce diverse (with high Distinct-N) yet coherent (with low PPL) text, avoiding the common pitfall of output collapse.

Our analysis, presented in Table \ref{tbl:detailed_performance_comparison}, shows that RCP-Merging excels in maintaining high output quality. It achieves an average Perplexity of 3.2, which is among the best of all methods, indicating that its outputs are highly fluent and linguistically sound. While some methods like DARE TIES AIM produce outputs with very high diversity of 80.7 average Distinct-N, this is often a symptom of degeneration, as confirmed by their high gibberish rates shown in Figure \ref{fig:gibberish_rate}. RCP-Merging, however, maintains a healthy diversity score of 47.0 without compromising coherence. These results suggest that our method successfully avoids output collapse and produces reliable, high-quality text, striking an effective balance between diversity and fluency.
\begin{table*}[ht]
	\centering

	\resizebox{0.95\textwidth}{!}{
	\begin{tabular}{l|cccc|cccc|cccc|cccc|cc}
		\toprule
		\multirow{3}{*}{\textbf{Method/Task}} & \multicolumn{4}{c|}{\textbf{Math}} & \multicolumn{4}{c|}{\textbf{Code}} & \multicolumn{4}{c|}{\textbf{BioMedicine}} & \multicolumn{4}{c|}{\textbf{Knowledge}} & \multicolumn{2}{c}{\textbf{Average}} \\
		\cmidrule(lr){2-5} \cmidrule(lr){6-9} \cmidrule(lr){10-13} \cmidrule(lr){14-17} \cmidrule(lr){18-19}
		
		& \multicolumn{2}{c|}{GSM8K} & \multicolumn{2}{c|}{AIME2024} & \multicolumn{2}{c|}{HumanEval} & \multicolumn{2}{c|}{LiveCodeBench} & \multicolumn{2}{c|}{PubMedQA} & \multicolumn{2}{c|}{MedQA} & \multicolumn{2}{c|}{ARC-C} & \multicolumn{2}{c|}{GPQA} & \multirow{2}{*}{D-N$\uparrow$} & \multirow{2}{*}{PPL$\downarrow$} \\
		\cmidrule(lr){2-3} \cmidrule(lr){4-5} \cmidrule(lr){6-7} \cmidrule(lr){8-9} \cmidrule(lr){10-11} \cmidrule(lr){12-13} \cmidrule(lr){14-15} \cmidrule(lr){16-17}
		
		& D-N$\uparrow$ & PPL$\downarrow$ & D-N$\uparrow$ & PPL$\downarrow$ & D-N$\uparrow$ & PPL$\downarrow$ & D-N$\uparrow$ & PPL$\downarrow$ & D-N$\uparrow$ & PPL$\downarrow$ & D-N$\uparrow$ & PPL$\downarrow$ & D-N$\uparrow$ & PPL$\downarrow$ & D-N$\uparrow$ & PPL$\downarrow$ & & \\
		\midrule
		
		\textbf{Base}& 59.9 & 8.8 & 51.0 & 1.5 & 71.2 & 6.5 & 48.5 & 5.1 & 69.6 & 19.3 & 50.3 & 4.5 & 33.2 & 6.7 & 54.3 & 2.6 & 54.8 & 6.9 \\
		\textbf{BioMedicine}& 65.4 & 10.8 & 70.7 & 1.7 & 62.6 & 8.9 & 19.8 & 31.5 & 65.4 & 18.1 & 64.9 & 3.1 & 61.5 & 2.9 & 62.4 & 10.8 & 59.1 & 11.0 \\
		\textbf{Reasoning}& 60.4 & 2.5& 59.1 & 1.4 & 54.3 & 1.6 & 56.6 & 1.6 & 75.5 & 3.9 & 67.6 & 1.8 & 57.5 & 2.2 & 52.5 & 1.5 & 60.4 & 2.1 \\
		\midrule
		Linear          & 42.0          & 4.9          & 6.9           & 1.7          & 9.8           & 3.4          & 26.2          & 7.6          & 38.5          & \underline{4.1} & 50.9          & 3.9          & 24.3          & 15.9         & 34.6          & \textbf{1.2} & 29.2          & 5.3 \\
		Task Arithmetic & 48.0          & 7.3          & 8.5           & 31.4         & 17.4          & 48.4         & 24.8          & \underline{1.7} & 47.2          & \textbf{3.8}    & 42.4          & 28.8         & 0.3           & 14.4         & 30.7          & \underline{1.3} & 27.4          & 17.1 \\
		TIES-Merging            & 24.1          & 5.1          & 11.7          & 7.5          & 13.6          & 3.0          & 2.1           & 3.8          & 54.4          & 6.7    & 27.3          & \textbf{1.3} & 4.6           & 11.6         & 1.8           & 10.6         & 17.5          & 6.2 \\
		DARE Linear     & 54.3          & 6.5          & 8.4           & 31.4         & 23.9          & 13.9         & 22.4          & 2.5          & 47.2          & \textbf{3.8}    & 49.7          & 5.3          & 0.4           & 14.4         & 30.7          & \underline{1.3} & 29.8          & 9.9 \\
		DARE TIES       & 44.9          & \textbf{2.0} & \textbf{97.7} & 22.7         & \textbf{48.0} & 26.3         & \textbf{96.2} & 9.9          & \underline{73.7} & 63.8   & \textbf{93.6} & 14.4         & \underline{95.6} & 26.3         & 33.2          & 11.9         & \underline{72.6} & 22.2 \\
		FuseLLM         & 44.6          & \textbf{2.0} & 65.8          & \underline{1.5} & 30.5          & 3.3          & 24.8          & 28.6         & 60.7          & 9.2    & 36.2          & 78.2         & 16.7          & \underline{3.2} & 12.4          & 21.0         & 36.5          & 18.4 \\
		DARE TIES \& AIM    & \textbf{78.0} & 63.7         & \underline{97.4} & 15.0         & 27.3          & 10.9         & \underline{95.9} & 19.9         & 58.1          & 31.5   & \underline{66.7} & 68.8         & \textbf{96.0} & 26.3         & \textbf{97.0} & 18.5         & \textbf{80.7} & 31.8 \\
		FuseLLM \& AIM     & 47.5          & 4.8          & 62.2          & 4.5          & 30.5          & 7.8          & 19.7          & 14.5         & 50.3          & 8.1    & 38.2          & 3.7          & 26.2          & 4.9          & 18.2          & \underline{1.3} & 36.8          & 6.2 \\
		Sens-Merging      & 29.0          & 2.8          & 10.3          & 1.8          & 38.3          & 2.0          & 51.1          & \textbf{1.5} & 7.3           & 5.5    & 10.0          & 8.3          & 19.1          & \textbf{3.1} & 11.5          & 2.1          & 22.1          & 3.3 \\
		CAT-Merging       & \underline{60.4} & \underline{2.5} & 10.7          & \textbf{1.4} & \underline{45.9} & \underline{1.9} & 51.7          & \textbf{1.5} & 28.3          & \textbf{3.8}    & 7.9           & \underline{2.5} & 20.2          & 3.9          & 10.6          & 2.1          & 29.5          & \textbf{2.5} \\
		\rowcolor{lightgray!25} \textbf{RCP-Merging} & 57.7 & 2.8 & 63.1 & \textbf{1.4} & 38.4 & \textbf{1.5} & 37.7 & 6.4 & \textbf{75.4} & 5.6 & 58.9 & 2.8 & 16.1 & 3.7 & \underline{51.5} & 1.4 & 47.0 & \underline{3.2} \\
		\bottomrule
	\end{tabular}}
	\caption{Distinct-N (D-N$\uparrow$) and PPL (PPL$\downarrow$) comparison when merging Qwen2.5-7B (Base), Meditron3-Qwen2.5-7B (BioMedicine) and DeepSeek-R1-Distill-Qwen-7B (Reasoning) on all datasets across Math, Code, BioMedicine, and Knowledge areas. For different merge methods, the best results within this subset are in \textbf{bold}, and the second-best are \underline{underlined}.}
    \label{tbl:detailed_performance_comparison}
\end{table*}

%% file: appendix/casestudy.tex
\label{app:case_study}
\begin{figure*}[t]
\centering
\includegraphics[width=1.0\textwidth]{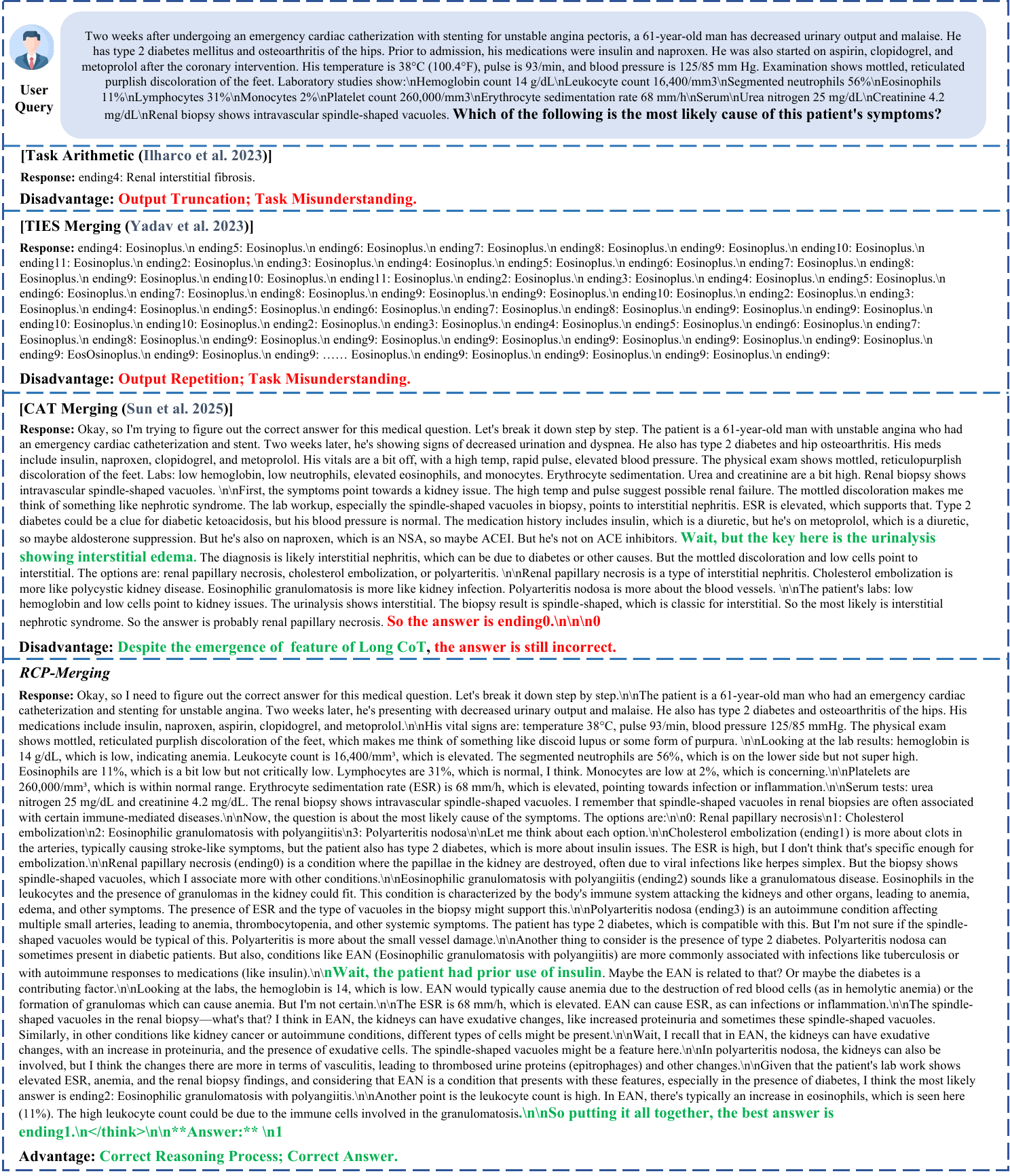}
\caption{Comparison of the features of model output using Task Arithmetic, TIES Merging, and CAT Merging with RCP-Merging on Qwen2.5-7B architecture merged model. The bold red font in the picture represents the disadvantage exposed in the previous model merging method, and the bold green font represents the advantages in this method.}
\label{fig:casestudy}
\end{figure*}

In this section, we provide a detailed analysis of the merged model output using different merge methods, as illustrated in Figure \ref{fig:casestudy}. 
The task is a complex medical diagnosis question about a 61-year-old man presenting with decreased urinary output and malaise two weeks after a cardiac catheterization. 
We compared the performance of several model merging techniques on the Qwen2.5-7B architecture.
The Task Arithmetic method misunderstood the task, leading to a truncated and incorrect response. 
Similarly, TIES-Merging also demonstrated task misunderstanding, which resulted in severe output repetition. 
An improvement was noted with CAT Merging, which generated a long CoT process; however, the reasoning was ultimately flawed and led to an incorrect answer. 
In contrast, our proposed RCP-Merging method demonstrated a correct reasoning process and arrived at the correct answer. 
It correctly analyzed the patient's symptoms, lab results, and biopsy findings, showcasing its superior performance on complex reasoning tasks.

%% file: appendix/generated_length.tex
\label{app:generated_length}
In the context of Large Reasoning Models, generated content length is also a key feature in distinguishing the performance of the model's reasoning capability. The longer the output content of the model is, the longer the thinking chain reasoning process, and the stronger self-thinking ability of the model can be reflected.

As shown in Table \ref{tab:generated_text_length}, RCP-Merging significantly surpasses other model merging methods on multiple benchmarks in terms of generated content length. Specifically, RCP-Merging generated the longest content on GSM8K, HumanEval, LiveCodeBench, ARC-C, and GPQA. It is noteworthy that the CAT method performs outstandingly on PubMedQA, MedQA, and AIME2024, achieving the highest average generation length. However, RCP-Merging's average generation length of 9921.1 is still highly competitive, far exceeding the Base model's 1213.3 and most other merging techniques. This result strongly indicates that RCP-Merging effectively integrates the deep reasoning capabilities of different expert models, thus enabling it to produce more detailed and complex chains of thought to solve problems.

\begin{table*}[ht]
	\centering

	\resizebox{0.98\textwidth}{!}{
	\begin{tabular}{l|cc|cc|cc|cc|c}
		\toprule
		\multirow{2}{*}{\textbf{Method/Task}} & \multicolumn{2}{c|}{\textbf{Math}} & \multicolumn{2}{c|}{\textbf{Code}} & \multicolumn{2}{c|}{\textbf{BioMedicine}} & \multicolumn{2}{c|}{\textbf{Knowledge}} & \multirow{2}{*}{\textbf{Average}} \\
		\cmidrule(lr){2-3} \cmidrule(lr){4-5} \cmidrule(lr){6-7} \cmidrule(lr){8-9}
		& GSM8K & AIME2024 & HumanEval& LiveCodeBench & PubMedQA & MedQA & ARC-C & GPQA & \\
		\midrule

		\textbf{Base}            & 416.0  & 4128.5  & 715.8   & 982.6   & 1743.3  & 33.4    & 903.5  & 783.1  & 1213.3  \\
		\textbf{BioMedicine}         & 297.5  & 1745.4  & 1236.7  & 5515.1  & 1867.3  & 1.5     & 232.0  & 1160.8 & 1507.0  \\
		\textbf{Reasoning}       & 1130.7 & 31946.2 & 13493.2 & 6878.0  & 3782.3  & 1549.1  & 3301.3 & 6643.2 & 8588.0  \\
		\midrule
		Linear          & 1206.6 & 1962.1  & 4546.5  & 4473.4  & 3826.3  & 811.5   & 1781.6 & 6071.8 & 3085.0  \\
		Task Arithmetic & 846.0  & 1644.5  & 2885.4  & 5534.2  & 3265.3  & 42.2    & 5164.3 & 6107.1 & 3186.1  \\
		TIES-Merging            & 1181.7 & 1313.3  & 4409.9  & 6983.1  & 3102.1  & 3495.3  & 3420.9 & 3913.3 & 3913.3  \\
		DARE Linear     & 770.1  & 1640.8  & 2346.8  & 5288.4  & 3267.6  & 1308.0  & 5159.7 & 6107.1 & 3236.1  \\
		DARE TIES       & 1416.9 & 672.9   & 1547.2  & 3356.8  & 2422.7  & 298.7   & 1076.4 & 3302.4 & 1761.8  \\
		FuseLLM         & 1487.0 & 21338.9 & 3117.6  & 5346.9  & 2810.9  & 1727.0  & 4439.4 & 5537.0 & 5725.3  \\
		DARE TIES \& AIM     & 953.9  & 1601.5  & 3729.9  & 994.9   & 1419.5  & 1057.9  & 1076.4 & 1134.5 & 1496.1  \\
		FuseLLM \& AIM     & 1125.7 & 15665.9 & 3992.3  & 4062.7  & 1477.5  & 2684.2  & 2924.0 & 4443.3 & 4547.0  \\
		Sens-Merging      & 4590.8 & 60721.3 & 12042.4 & 12170.4 & 5481.2  & 11356.5 & 4804.9 & 5568.9 & 14592.1 \\
		CAT-Merging       & 1130.7 & \textbf{71342.4} & 15485.4 & 12244.0 & \textbf{26947.1} & \textbf{12876.1} & 4134.7 & 5888.8 & \textbf{18750.2} \\
		\rowcolor{lightgray!25} \textbf{RCP-Merging} & \textbf{4905.1} & 18766.8 & \textbf{17662.0} & \textbf{18446.5} & 3048.0 & 4757.5 & \textbf{5209.5} & \textbf{6573.2} & 9921.1 \\
		\bottomrule
	\end{tabular}}
    	\caption{Comparison of the generated content token length between RCP-Merging and other merge methods on the Qwen2.5-7B base model across eight benchmarks, with the average token length calculated. The longest generated text length under a certain dataset is highlighted in \textbf{bold}.}
        \label{tab:generated_text_length}
\end{table*}





%% file: appendix/model_size.tex
\label{app:model_size}
To assess the scalability of our approach, we apply RCP-Merging to a smaller model series, using the Qwen-1.5B architecture. The experiment involves the Qwen2.5-1.5B as Base model \citep{Yang24qwen},
using BioQwen-1.5B \citep{li2025bioqwen} as BioMedicine model
and the DeepSeek-R1-Distill-Qwen-1.5B Reasoning model.
Drawing upon foundational research into the behavior of neural language models, we can infer the relationship between model scale and the regularization required for complex reasoning tasks, which is also addressed by our algorithm. 
Previous work by \citet{kaplan2020scaling,chen2025understanding} established that model performance scales predictably with size, implying smoother training dynamics for larger models. These findings also suggested that smaller models exhibit sharper and more rugged loss landscapes. 
This characteristic of smaller models may have a greater magnitude and variance, a phenomenon. 
Consequently, in the context of the conflict metric in Equation \ref{eq:constraint_metric}, the pre-factor $C^t_{i,k}$ is expected to be substantially larger for a smaller model, as it is directly influenced by these gradient fluctuations. 
To counterbalance this inherent instability and guide the model towards a robust reasoning optimum, it becomes necessary to apply a stronger regularization penalty. 
Therefore, a smaller model necessitates a larger value for the regularization coefficient $\lambda$ to adequately constrain its optimization trajectory. 
Therefore, when handling a 1.5B model, we suggest using $\lambda = 0.7$ as the hyperparameter to determine the intensity of the reasoning constraint in RCP-Merging.

As shown in Table \ref{tab:qwen1.5B_performance}, even at a smaller scale, RCP-Merging continues to demonstrate its superiority. 
It achieves the highest average score of 47.8, leading in performance on the Math, BioMedicine, and Knowledge benchmarks. 
While another method, CAT Merging, shows slightly better performance on the Code benchmark, RCP-Merging's overall performance across all tasks is dominant. 
This result indicates that our method is not reliant on large model sizes and can be effectively applied to more compact and efficient models.

\begin{table}[H]
	\centering


	\resizebox{0.98\columnwidth}{!}{
	\begin{tabular}{l|cccc|c}
		\toprule
		\textbf{Method/Task} & \textbf{Math} & \textbf{Code} & \textbf{BioMedicine} & \textbf{Knowledge} & \textbf{Average} \\
		\midrule

		\textbf{Base}                & 33.3 & 35.4 & 7.5 & 36.8 & 28.3 \\
		\textbf{BioMedicine}           & 24.0 & 24.4 & 46.5 & 35.7 & 32.7 \\
		\textbf{Reasoning}         & 56.7 & 43.3 & 26.5 & 48.8 & 43.8 \\
        \midrule
		Linear          & 23.7 & 21.3 & 7.5 & 26.1 & 19.7 \\
		Task Arithmetic & 22.1 & 25.0 & 16.5 & 25.6 & 22.3 \\
		TIES-Merging            & 22.8 & 29.3 & 18.0 & 33.3 & 25.9 \\
		DARE Linear     & 22.7 & 23.8 & 13.0 & 25.6 & 21.3 \\
		DARE TIES       & 23.6 & 26.2 & 16.0 & 30.0 & 24.0 \\
		FuseLLM         & 24.1 & 22.0 & 8.0 & 24.3 & 19.6 \\
		DARE TIES \& AIM   & 28.1 & 26.2 & 14.5 & 38.6 & 26.9 \\
		FuseLLM \& AIM     & 29.0 & 28.0 & 22.0 & 32.6 & 27.9 \\
		Sens-Merging      & 40.4 & 31.1 & 30.0 & 41.4 & 35.7 \\
		CAT-Merging       & 44.1 & \textbf{44.5} & 21.0 & 39.3 & 37.2 \\
		\rowcolor{lightgray!25} \textbf{RCP-Merging}  & \textbf{54.4} & 38.4 & \textbf{46.5} & \textbf{51.7} & \textbf{47.8} \\
		\bottomrule
	\end{tabular}}
    	\caption{Performance comparison of merging Qwen2.5-1.5B (Base),  BioQwen-1.5 (BioMedicine), and DeepSeek-R1-Distill-Qwen-1.5B (Reasoning) on four datasets across Math, Code, BioMedicine, and Knowledge areas. The best performance under a certain dataset is highlighted in \textbf{bold}.}
        \label{tab:qwen1.5B_performance}
\end{table}